\documentclass[10pt,twocolumn,letterpaper]{article}

 \usepackage[pagenumbers]{cvpr} 

\usepackage[svgnames,table,rgb]{xcolor} 
\definecolor{cvprblue}{rgb}{0.21,0.49,0.74}
\usepackage[pagebackref,breaklinks,colorlinks,allcolors=cvprblue]{hyperref}

\usepackage{hyperref}
\usepackage{url}
\usepackage{algorithm}
\usepackage{algorithmicx}
\usepackage{amsmath}
\usepackage{booktabs} 
\usepackage{array}
\definecolor{mylightgray}{rgb}{0.94,0.94,0.94}
\definecolor{darkred}{rgb}{0.9, 0, 0}
\usepackage{algpseudocode}
\usepackage{algcompatible} 
\usepackage{color}
\usepackage[symbol]{footmisc}
\usepackage{graphicx}
\usepackage{wrapfig}
\usepackage{amssymb}
\algtext*{ENDIF}
\algtext*{ENDFOR}

\def\paperID{15457} 
\def\confName{CVPR}
\def\confYear{2025}

\title{Decoupled Video Generation with Chain of Training-free Diffusion Model Experts}

\author{
    Wenhao Li\textsuperscript{1}\thanks{Equal contribution}, 
    Yichao Cao\textsuperscript{2}\footnotemark[1], 
    Xiu Su\textsuperscript{3}\thanks{Corresponding author}, 
    Xi Lin\textsuperscript{4}, 
    Shan You\textsuperscript{5}, 
    Mingkai Zheng\textsuperscript{1}, 
    Yi Chen\textsuperscript{6}, 
    Chang Xu\textsuperscript{1}
}

\begin{document}
\twocolumn[
\maketitle

\vspace{-27pt}

\begin{center}
\textsuperscript{1}University of Sydney, 
\textsuperscript{2}Southeast University, 
\textsuperscript{3}Central South University, 
\textsuperscript{4}Shanghai Jiaotong University, 
\textsuperscript{5}Sensetime Research, 
\textsuperscript{6}Hong Kong University of Science and Technology
\end{center}
\vspace{0.5cm}
]

\begin{abstract}
Video generation models hold substantial potential in areas such as filmmaking. However, current video diffusion models need high computational costs and produce suboptimal results due to extreme complexity of video generation task. In this paper, we propose \textbf{ConFiner}, an efficient video generation framework that decouples video generation into easier subtasks: structure \textbf{con}trol and spatial-temporal re\textbf{fine}ment. It can generate high-quality videos with chain of off-the-shelf diffusion model experts, each expert responsible for a decoupled subtask. During the refinement, we introduce coordinated denoising, which can merge multiple diffusion experts' capabilities into a single sampling. Furthermore, we design ConFiner-Long framework, which can generate long coherent video with three constraint strategies on ConFiner. Experimental results indicate that with only 10\% of the inference cost, our ConFiner surpasses representative models like Lavie and Modelscope across all objective and subjective metrics. And ConFiner-Long can generate high-quality and coherent videos with up to 600 frames. All the code will be available at project website: \url{https://confiner2025.github.io}.
\end{abstract}    
\section{Introduction}
\label{sec:intro}
Generative AI \cite{b1,b2,b3} has recently emerged as a hotspot in research, influencing various aspects of our daily life. For visual AIGC, numerous image generation models, such as Stable Diffusion \cite{b4} and Imagen \cite{b5}, have achieved significant success. These models can create high-resolution images that are rich in creativity and imagination, rivaling those created by human artists. Compared to image generation, video generation models \cite{b6,b7,b8,b9} hold higher practical value with the potential to reduce expenses in the fields of filmmaking and animation.
\begin{figure}[t!]
  \centering
\includegraphics[width=0.48\textwidth]{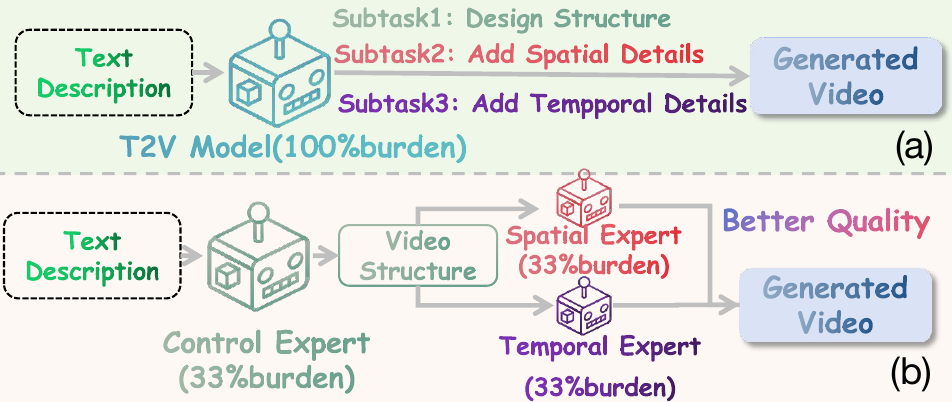}
  \caption{(a) Conventional video generation process. (b) Motivation of the proposed ConFiner.}
  \vspace{-15pt}
  \label{coreidea}
\end{figure}

However, current video generation models are still in their early stages of development. Existing video diffusion models can primarily be categorized into three types. The first type uses T2I (Text to Image) models to generate videos directly without further training \cite{b33,b56,b57,b58}. The second type incorporates a temporal module into T2I models and trains on video datasets \cite{b10,b11,b40}. The third type is trained from scratch \cite{b46,b59,b60,b61}. Regardless of which type, these methods use a single model to undertake the entire task of video generation, like \cref{coreidea}(a). However, video generation is extremely intricate \cite{b34}. After our in-depth analysis, we believe that this complex task consists of three subtasks: modeling the \textit{video structure}, which includes designing the overall visual structure and plot; generating \textit{spatial details}, ensuring each frame with sufficient clarity and high aesthetic score; and producing \textit{temporal details}, maintaining consistency and coherence between frames to ensure natural and logical transitions. Therefore, relying on a single model to handle such a complex and multidimensional task is challenging.

\begin{figure*}[t!]
  \centering  \includegraphics[width=\textwidth]{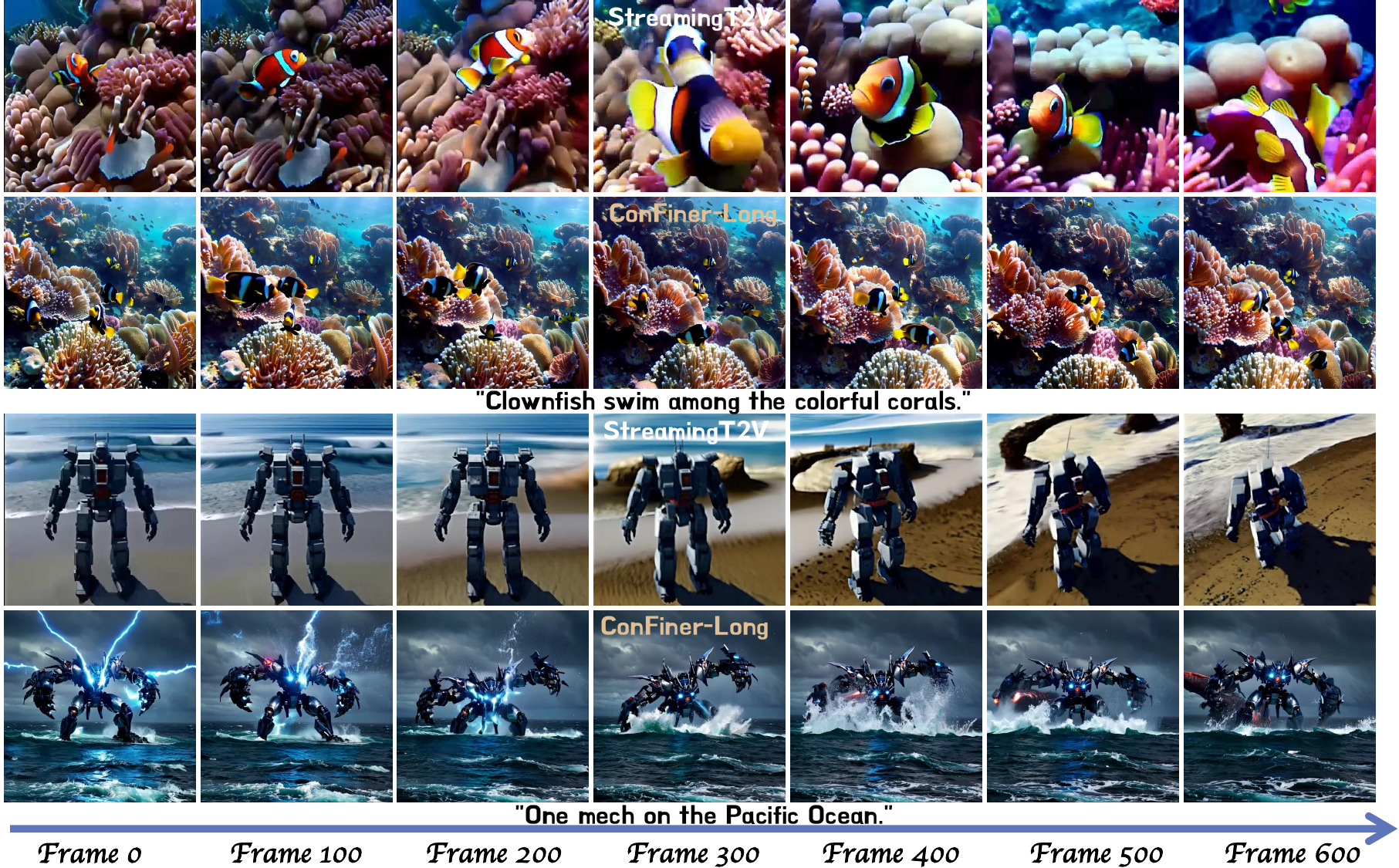}
  \caption{\textbf{Comparison between Our ConFiner-Long and StreamingT2V \cite{b54}.} We exhibit better consistency and imaging quality.}
  \vspace{-10pt}
  \label{img1}
\end{figure*}
Overall, there are three main challenges in the field of video generation \cite{b62,b63,b64,b65}: \textit{i}) The quality of the generated videos is low, hard to achieve high-quality temporal and spatial modeling simultaneously \cite{b34}. \textit{ii}) The generation process is time-consuming, often requiring hundreds of inference steps \cite{b66}. Utilizing a single model to handle complex video generation task is one of the key reasons for these two issues. \textit{iii}) The length of the generated videos are typically short \cite{b67}. Due to limitations in VRAM, the length of videos generated in a single attempt generally ranges between only 2-3 seconds.

In order to enhance generation quality, some methods employ multiple models on different resolutions or in different spaces to perform progressive generation. Some methods \cite{b47,b53,b12,b46} train several diffusion models on gradually increasing resolutions to first generate low-resolution videos, and then progressively scale up. Show-1 \cite{b48} trains a model in pixel space to generate low-quality videos, followed by a latent space model to enhance quality. Compared to methods using a single model, these approaches achieve higher performance. However, each model still needs to handle both spatial and temporal modeling. This leaves each model still heavily burdened.

To improve quality of videos while reducing inference time, we rethink the demands of video generation tasks, which include modeling video structure, generating spatial details, and producing temporal details. We find out that a more rational approach is utilizing three specialized models, each handling one demand. By doing so, these models can collaboratively accomplish this comprehensive task. To this end, we propose a framework named ConFiner, which decouples the video generation process into three parts: structure control, temporal refinement, and spatial refinement. During generation, we employ chain of three ready-made diffusion experts, each specializing in respective tasks, like \cref{coreidea}(b). In the control stage, a highly controllable T2V (Text to Video) model is employed as control expert, tasked with structure control. During the refinement stage, a T2I model and a T2V model skilled at generating details are employed as spatial and temporal experts to refine details. This framework can reduce the burden on individual models, enhancing both the quality and speed of generation. Moreover, as it utilizes ready-made diffusion experts, this framework does not incur additional training costs.



\begin{figure*}[t!]
  \centering
\includegraphics[width=\textwidth]{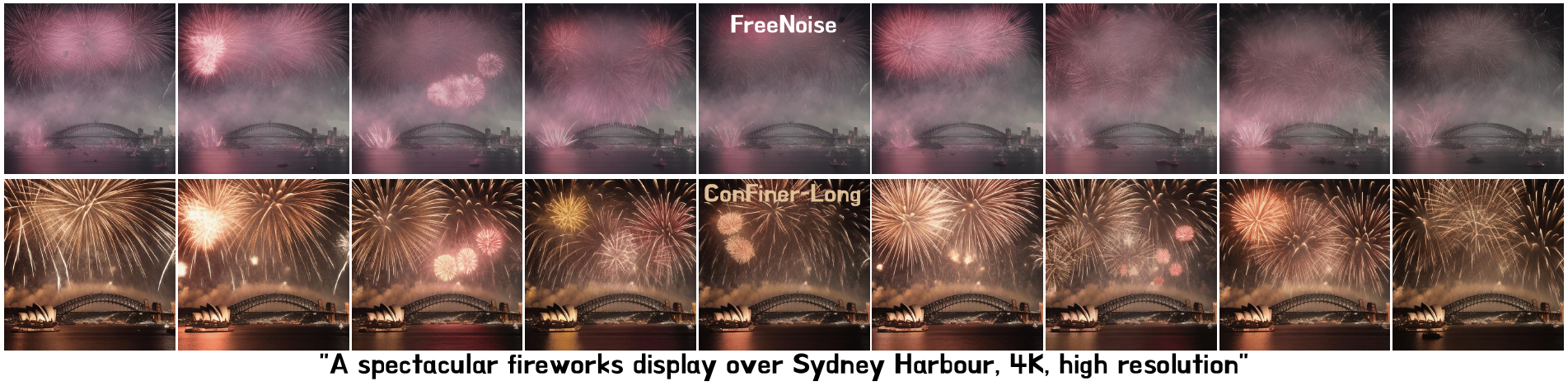}
  \caption{\textbf{Comparison of Our ConFiner-Long with FreeNoise \cite{b55}.} We achieve much better imaging clarity and quality.}
  \vspace{-15pt}
  \label{comparison}
\end{figure*}

Furthermore, based on ConFiner, we propose ConFiner-Long framework, which can generate long videos by ensuring the coherence and consistency between video segments. As the initial noise significantly impacts the final videos, we first introduce a segments consistency initialization strategy to ensure the consistency of the initial noise between segments by sharing a base noise. Additionally, in order to enhance the coherence of the motion between segments, we propose a coherence guidance strategy that uses the gradient of noise differences between two segments to guide the denoising direction. Also, to address the flickering problem at the junctions of segments, we design a staggered refinement strategy that staggers the control stage and the refinement stage. It places the tail of one video structure and the head of the next into the same refinement process to achieve more natural transitions between segments.

Experimental results have shown that ConFiner requires only 9 sampling steps (less than 5 seconds) to surpass the performance of models like AnimateDiff-Lightning \cite{b40}, LaVie \cite{b12}, and ModelScope T2V \cite{b10} with 100-step sampling (more than 1 minute). Furthermore, ConFiner-Long can generate high-quality coherent videos up to 600 frames long. To sum up, our contributions are as follows:
\begin{enumerate}
    \item We introduced ConFiner, which decouples the video generation task into three sub-tasks. It utilizes three ready-made diffusion experts, each handling its specialized task. This approach reduces the model's burden, enhancing the quality and speed of generation.
    \item We designed coordinated denoising strategy, allowing two experts on different noise schedulers to collaborate timestep-wise in video generation process.
    \item We proposed ConFiner-Long framework, which harmonizes the initial states, generation directions, and transitions between segments to achieve high-quality, coherent long video production.
\end{enumerate}
\section{Related Work}
\textbf{Diffusion models (DMs).} DMs have achieved remarkable successes in the generation of images \cite{b24,b68,b69,b70,b71,b72}, music \cite{b25,b73,b74,b75,b76}, and 3D models \cite{b28,b29,b77,b78,b79,b80}. These models typically involve thousands of timesteps, with a scheduler that manages the noise level. Diffusion models consist of two processes \cite{b31}. In the forward process, noise is progressively added to the original data until it is completely transformed into noise. During the reverse denoising process, the model starts with random noise and gradually eliminates the noise using a denoising model, ultimately transforming it into a target sample.

\noindent{\textbf{Video Diffusion Models (VDMs).}} Compared to the success of diffusion models in areas like image generation, VDMs are still at a very early stage. Some methods \cite{b33,b56,b57,b58} use stable diffusion without additional training for direct video generation. These methods suffer from poor coherence and evident visual tearing. Some Models \cite{b10,b11,b40} convert the U-Net of stable diffusion \cite{b12} into a 3D U-Net through the addition of temporal convolution or attention, and train it on video datasets to achieve video generation. And some other methods \cite{b46,b59,b60,b61} are trained from scratch. However, the generation quality and speed of these methods are unsatisfactory, which we attribute to the overwhelming burden placed on a single model to handle the complexity of video generation tasks.
\section{Method}
\subsection{Overview}
Our ConFiner consists of two stages: the control stage and the refinement stage. In the control stage, it generates a video structure containing coarse-grained spatio-temporal information, which determines the overall structure and plot of the final video. During the refinement stage, it refines spatial and temporal details based on video structure. In this stage, we propose coordinated denoising to enable cooperation of spatial expert and temporal expert. Based on ConFiner, we introduce ConFiner-Long framework for producing coherent and consistent long videos.

\subsection{Revisiting Diffusion Models}
The workflow of diffusion models consists of two processes: the forward process and the reverse denoising process. The forward process from timestep 0 to timestep $t$ can be expressed as follows:
\begin{equation}
\mathbf{x}_t = \sqrt{\overline{\alpha}_t} \mathbf{x}_0 + \sqrt{1 - \overline{\alpha}_t} \epsilon\label{eq1}
\end{equation}
where $\alpha_t = 1-\beta_t$, $\overline{\alpha}_t = \prod_{i=1}^t \alpha_i$, $t$ is the diffusion step, $\mathbf{\epsilon}$ is a random noise sampled from Standard Gaussian Distribution $\mathcal{N}(0,1)$ and $\beta_t$ is a small positive constant between 0 and 1, representing the noise level of each timestep.

\begin{figure*}[t!]
  \centering
  \includegraphics[width=\textwidth]{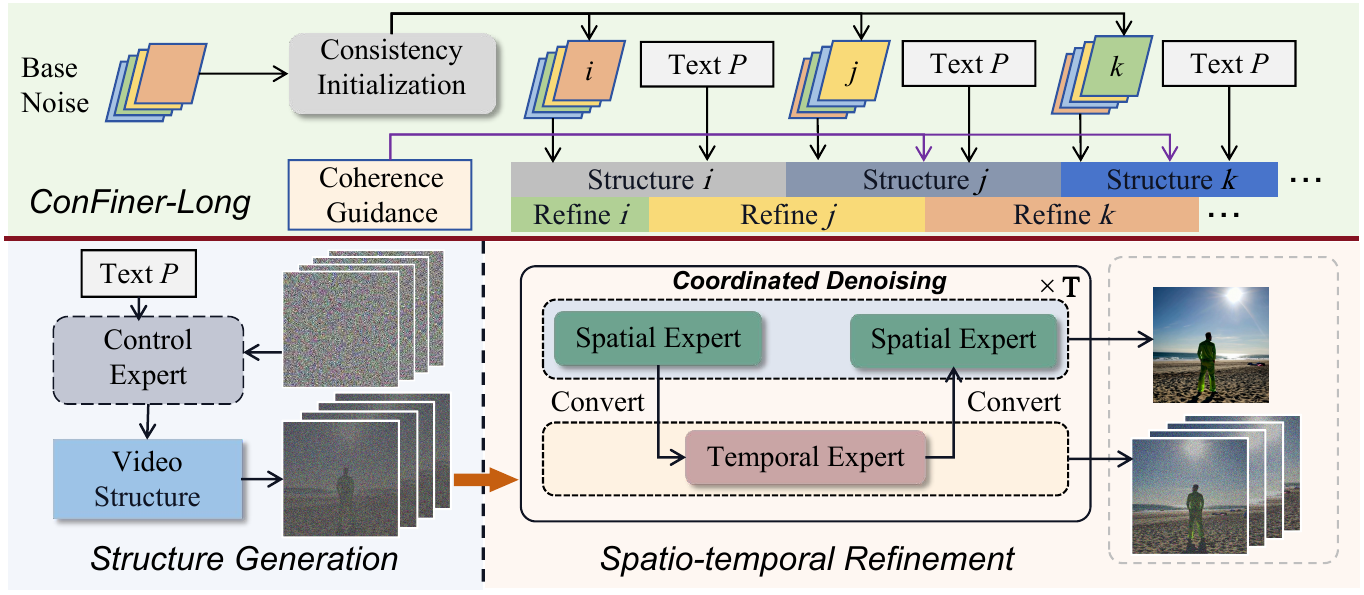}
  \caption{\textbf{Pipeline of Our ConFiner and ConFiner-Long.} ConFiner decouples the video generation process. Firstly, control expert generates a video structure. Subsequently, temporal and spatial experts perform the refinement of spatio-temporal details. Spatial and temporal experts work together with our coordinated denoising. By adding consistency initialization, coherence guidance and staggered refinement to ConFiner, ConFiner-Long can generate coherent long videos.}
  \vspace{-10pt}
  \label{main_fig}
\end{figure*}

During the reverse denoising process, starting from a random noise at timestep $T$, the denoising model progressively predicts $\mathbf{x}_{t-1}$ from $\mathbf{x}_t$, ultimately getting the target data $\mathbf{x}_0$. Taking DDIM \cite{b35} as an example, the denoising model initially uses $\mathbf{x}_t$ to predict the noise. Then, $\mathbf{x}_t$ and the predicted noise are utilized together to predict $\mathbf{x}_0$ via the following expression:
\begin{equation}
\hat{\mathbf{x}}_0 = \frac{\mathbf{x}_t - \sqrt{1 - \overline{\alpha}_t} \epsilon^{(t)}_{\theta}(\mathbf{x}_t)}{\sqrt{\overline{\alpha}_t}}
\label{eq2}
\end{equation}
where $\epsilon^{(t)}_{\theta}(x_t)$, $\hat{\mathbf{x}}_0$ represents the predicted noise and $\mathbf{x}_0$.

Then, based on the predicted noise and $\hat{\mathbf{x}}_0$ , a prediction for $\mathbf{x}_{t-1}$ is derived as:
\begin{equation}
\mathbf{x}_{t-1} = \sqrt{\overline{\alpha}_{t-1}} \cdot \hat{\mathbf{x}}_0 + \sqrt{1-\overline{\alpha}_{t-1} \cdot }\epsilon^{(t)}_{\theta}(\mathbf{x}_t) \label{eq3}
\end{equation}

By combining \cref{eq2} and \cref{eq3}, single-step denoising can be expressed as:

\begin{equation}
\hat{\mathbf{x}}_0, \mathbf{x}_{t-1} = \text{Denoising}(\theta, \mathbf{x}_t, t, S)\label{eq4}
\end{equation}
where $S$ denotes the noise scheduler and $\theta$ represents the corresponding denoising model.

\subsection{Video Structure Generation}
In the control stage, we select a video diffusion model skilled at handling video structure and employ it as control expert. The scheduler used in this expert can be denoted as $S_\text{con}$. During inference, to reduce computational overhead, we opt for a DDIM scheduler with a total inference step of $T_{i_1}$. When conducting inference, the list of timesteps utilized is: $[t_1(i_1), t_2(i_1),..., t_{T_{i_1}}(i_1)]$. The selection of timesteps is made at uniform intervals.

After obtaining the timesteps list, we start with a random noise $\mathcal{V}_{t_{T_{i_1}(i_1)}}$ and progressively denoise over these timesteps, getting the first version of the video $\mathcal{V}_0$. Single-step sampling from \cref{eq4} can be rewritten as follows.
\begin{equation}
\hat{\mathcal{V}}_0(t_k(i_1)), \mathcal{V}_{t_{k-1}(i_1)} = \text{Denoising}(\theta_\text{con}, \mathcal{V}_{t_{k}(i_1)}, t_k(i_1), S_\text{con})\label{eq6}
\end{equation}
where $\hat{\mathcal{V}}_0(t_k(i_1))$ represents the predicted $\mathcal{V}_0$ at timestep $t_k(i_1)$, $\mathcal{V}_{t_{k}(i_1)}$ denotes $\mathcal{V}$ at timestep $t_k(i_1)$, $\theta_\text{con}$ represents control expert and $S_\text{con}$ is the scheduler of control expert.

While we completed the entire sampling to obtain the first version of video $\mathcal{V}_0$, the quality and coherence of the video are compromised due to our choice of a small $T_{i_1}$. 

Therefore, we introduce $T_e$ steps of noise to $\mathcal{V}_0$. This operation is intended to create refinement opportunities for spatial and temporal experts. In this noise addition process, we utilize the Scheduler $S_{s}$ from the spatial expert used in refinement stage, resulting in the noisy video $\mathcal{V}'_{T_e}$ at timestep $T_e$. Transformed from \cref{eq1}, this noise addition process can be expressed as:
\begin{equation}
\mathcal{V}'_{T_e} = \sqrt{\overline{\alpha}_{T_e}(S_{s})}  \cdot \mathcal{V}_0 + \sqrt{1 - \overline{\alpha}_{T_e}(S_{s})} \cdot  \epsilon\label{eq7}
\end{equation}
where $\overline{\alpha}_{T_e}(S_{s})$ is the $\overline{\alpha}_t$ in scheduler $S_s$ at timestep $T_e$.

\subsection{Spatial and Temporal Details Refinement}
During the refinement stage, we add spatial and temporal details with spatial expert and temporal expert in the process of transforming $\mathcal{V}'_{T_e}$ to $\mathcal{V}'_0$. Similar to the control stage, we select $T_{i_2}$ steps for sampling between timestep $T_e$ and timestep 0. The list of timesteps used is: $[t_1(i_2), t_2(i_2),..., t_{T_{i_2}}(i_2)]$.

Given that two experts respectively excel in spatial and temporal modeling, we aim to synergistically utilize both experts in the process of denoising $\mathcal{V}'_{T_e}$ to $\mathcal{V}'_0$, thus enhancing the spatio-temporal detail. A straightforward approach is alternating between the two experts at each timestep, leveraging the strengths of both models concurrently. In this case, \cref{eq4} can be rewritten as follows:
\begin{equation}
\begin{aligned}
\hat{\mathcal{V}}_0(t_k(i_2)), \mathcal{V}'_{t_{k-1}(i_2)} &= \text{Denoising}(\theta_{X}, \mathcal{V}'_{t_{k}(i_2)}, t_k(i_2), S_{s}) \\
\text{where } \theta_{X} &= 
\begin{cases} 
\theta_{S} & \text{if } k \equiv 2 \pmod{0} \\
\theta_{T} & \text{if } k \equiv 2 \pmod{1}
\end{cases}
\end{aligned}
\label{eq9}
\end{equation}
where $\theta_{S}$, $\theta_{T}$ represent sptial expert and temporal expert, and $S_{s}$ denotes spatial expert's scheduler.

However, this method is ineffective because spatial expert and temporal expert are often on different noise scheduler. The data distributions for the spatial and temporal experts at the same timestep are inconsistent. The original data is on the scheduler of spatial expert, and directly switching to the scheduler of temporal expert at a certain timestep leads to conflicts and inconsistencies. To transform $\mathcal{V}'_{t_{k}(i_2)}$ to $\mathcal{V}'_{t_{k-1}(i_2)}$, we provide two options.

\paragraph{Option 1 (Standard Denoising):}Since the original data $\mathcal{V}'_{T_e}$ is on the scheduler of spatial expert, we can directly employ the spatial expert for denoising at time step $t_k(i_2)$:
\begin{equation}
\hat{\mathcal{V}}_0(t_k(i_2)), \mathcal{V}'_{t_{k-1}(i_2)} = \text{Denoising}(\theta_{S}, \mathcal{V}'_{t_{k}(i_2)}, t_k(i_2), S_{s})\label{eq10}
\end{equation}

\paragraph{Option 2 (Coordinated Denoising):} Although two experts' schedulers differ, both schedulers share the same distribution at timestep 0. Hence, we can utilize timestep 0 to establish a connection between the two schedulers, facilitating the concurrent use of two experts within the same timestep. The specific details of this process are as follows.

First, at timestep $t_k(i_2)$, given $\mathcal{V}'_{t_{k}(i_2)}$, we employ the spatial expert for a one-step inference as \cref{eq10}. After obtaining the predicted $\hat{\mathcal{V}}_0(t_k(i_2))$, it can be converted to $\mathcal{V}''_{t_k(i_2)}$ on the scheduler of temporal expert.
\begin{equation}
\mathcal{V}''_{t_k(i_2)} = \sqrt{\overline{\alpha}_{t_k(i_2)}(S_{t})}  \cdot \hat{\mathcal{V}}_0(t_k(i_2)) + \sqrt{1 - \overline{\alpha}_{t_k(i_2)}(S_{t})} \cdot  \epsilon \label{eq11}
\end{equation}
where $S_t$ represents the noise scheduler of temporal expert.

\begin{algorithm}[t!]
\caption{ConFiner (Control + Refinement)}
\begin{algorithmic}[1]
\STATE \textbf{Input:} Prompt $P$, Control Expert $Con$, Spatial Expert $S$, Temporal Expert $T$, Noisy timestep $T_e$
\STATE \textbf{Output:} Generated video $\mathcal{V}$
\STATE $\mathcal{V}_0 \leftarrow \text{Generate}(P,Con)$ \hfill $\triangleright$ Generate coarse video.
\STATE $Video\_Structure \leftarrow \text{Add noise}(\mathcal{V}_0,T_e,Con)$ 
\STATE $\mathcal{V}_{T_e}' \leftarrow Video\_Structure$ \hfill $\triangleright$ Extract video structure.
\FOR{each refinement step $T_k$}
    \IF{Standard Denoising}
        \STATE $\mathcal{V}_{T_{k-1}}' \leftarrow \text{Denoise}(\mathcal{V}_{T_k}',T_k,S)$ 
    \ELSIF{Coordinated Denoising}
        \STATE $\mathcal{V}'_0(T_k) \leftarrow \text{Denoise}(\mathcal{V}_{T_k}',T_k,S,P)$  
        \STATE $\mathcal{V}''_{T_k} \leftarrow \text{Add noise}(\mathcal{V}'_0(T_k),T_k,T)$ 
        \STATE $\mathcal{V}''_0(T_{k}) \leftarrow \text{Denoise}(\mathcal{V}_{T_k}'',T_k,T,P)$ 
        \STATE $\mathcal{V}'_{T_{k}} \leftarrow \text{Add noise}(\mathcal{V}''_0(T_{k}),T_{k},S)$  
        \STATE $\mathcal{V}_{T_{k-1}}' \leftarrow \text{Denoise}(\mathcal{V}_{T_k}',T_k,S,P)$ 
    \ENDIF
\ENDFOR
\STATE \Return \textbf{Return} $\mathcal{V}=\mathcal{V}'_0$, with refined spatial-temporal details.
\end{algorithmic}
\end{algorithm}

Then we can employ the temporal expert for denoising:
\begin{equation}
\hat{\mathcal{V}}_0(t_{k}(i_2)), \mathcal{V}''_{t_{k-1}(i_2)} = \text{Denoising}(\theta_{T}, \mathcal{V}''_{t_{k}(i_2)}, t_k(i_2), S_{t}) \label{eq12}
\end{equation}

This version of $\hat{\mathcal{V}}_0(t_{k}(i_2))$ predicted by temporal expert contains richer temporal information and demonstrates enhanced inter-frame coherence. Subsequently, we transform $\hat{\mathcal{V}}_0(t_{k}(i_2))$ using the scheduler of spatial expert into a $\mathcal{V}'_{t_k(i_2)}$ with more extensive temporal information.
\begin{equation}
\mathcal{V}'_{t_k(i_2)} = \sqrt{\overline{\alpha}_{t_k(i_2)}(S_{s})}  \cdot \hat{\mathcal{V}}_0(t_{k}(i_2)) + \sqrt{1 - \overline{\alpha}_{t_k(i_2)}(S_{s})} \cdot  \epsilon \label{eq13}
\end{equation}

Finally, the spatial expert is used again to predict \(\mathcal{V}'_{t_{k-1}(i_2)}\) including more spatio-temporal details as \cref{eq10}.

\begin{table*}[t!]
\centering
\renewcommand{\arraystretch}{1}
\small
\setlength{\tabcolsep}{11pt} 
\begin{tabular}{cccccccc}
\toprule
Method & \shortstack{Inference \\ Steps} & \shortstack{Subject \\ Consistency}$\uparrow$ & \shortstack{Motion \\ Smoothness}$\uparrow$ & \shortstack{Aesthetic \\ Quality}$\uparrow$ & \shortstack{Imaging \\ Quality}$\uparrow$ \\ 
\midrule
Lavie \cite{b12} & 10 & $0.940 \pm 0.001$ & $0.967 \pm 0.001$ & $0.570 \pm 0.003$ & $0.658 \pm 0.008$ \\
Lavie \cite{b12} & 20 & $0.954 \pm 0.002$ & $0.966 \pm 0.002$ & $0.587 \pm 0.001$ & $0.683 \pm 0.001$ \\
Lavie \cite{b12} & 50 & $0.958 \pm 0.004$ & $0.965 \pm 0.006$ & $0.597 \pm 0.005$ & $0.696 \pm 0.005$ \\
Lavie \cite{b12} & 100 & $0.957 \pm 0.003$ & $0.965 \pm 0.001$ & $0.596 \pm 0.006$ & $0.695 \pm 0.007$ \\
\midrule
AnimateDiff-Lightning \cite{b40} & 10 & $0.983 \pm 0.002$ & $0.983 \pm 0.001$ & $0.635 \pm 0.002$ & $0.689 \pm 0.001$ \\
AnimateDiff-Lightning \cite{b40} & 20 & $0.984 \pm 0.004$ & $0.980 \pm 0.002$ & $0.636 \pm 0.006$ & $0.697 \pm 0.002$ \\
AnimateDiff-Lightning \cite{b40} & 50 & $0.981 \pm 0.004$ & $0.971 \pm 0.003$ & $0.638 \pm 0.002$ & $0.705 \pm 0.003$ \\
AnimateDiff-Lightning \cite{b40} & 100 & $0.977 \pm 0.004$ & $0.964 \pm 0.003$ & $0.623 \pm 0.006$ & $0.699 \pm 0.005$ \\
\midrule
Modelscope T2V \cite{b10} & 10 & $0.983 \pm 0.002$ & $0.980 \pm 0.001$ & $0.570 \pm 0.002$ & $0.670 \pm 0.004$ \\
Modelscope T2V \cite{b10} & 20 & $0.985 \pm 0.004$ & $0.980 \pm 0.003$ & $0.575 \pm 0.003$ & $0.702 \pm 0.004$ \\
Modelscope T2V \cite{b10} & 50 & $0.988 \pm 0.002$ & $0.990 \pm 0.001$ & $0.592 \pm 0.002$ & $0.716 \pm 0.002$ \\
Modelscope T2V \cite{b10} & 100 & $0.987 \pm 0.002$ & $0.990 \pm 0.000$ & $0.594 \pm 0.001$ & $0.715 \pm 0.004$ \\
\midrule
\rowcolor{mylightgray}
\textbf{ConFiner} w/ Lavie & 9 & $0.993 \pm 0.000$ & $0.991 \pm 0.000$ & $0.699 \pm 0.005$ & $0.734 \pm 0.005$ \\
\rowcolor{mylightgray}
\textbf{ConFiner} w/ Lavie & 18 & $0.993 \pm 0.002$ & $0.990 \pm 0.001$ & $0.703 \pm 0.009$ & $0.739 \pm 0.004$ \\
\rowcolor{mylightgray}
\textbf{ConFiner} w/ Modelscope & 9 & $0.994 \pm 0.000$ & $0.991 \pm 0.000$ & $0.698 \pm 0.006$ & $0.731 \pm 0.003$ \\
\rowcolor{mylightgray}
\textbf{ConFiner} w/ Modelscope & 18 & $\mathbf{0.994} \pm 0.002$ & $\mathbf{0.991} \pm 0.002$ & $\mathbf{0.707} \pm 0.004$ & $\mathbf{0.739} \pm 0.004$ \\
\bottomrule
\end{tabular}
\vspace{-5pt}
\caption{\textbf{Objective Evaluation Results.} In this experiment, ConFiner utilized AnimateDiff-Lightning as the control expert and selected stable diffusion 1.5 for spatial expert. Lavie and Modelscope T2V are chosen as temporal expert.}
\vspace{-5pt}
\label{tab:objective}
\end{table*}

\begin{table*}[t!]
\centering
\renewcommand{\arraystretch}{1.05} 
\small
\setlength{\tabcolsep}{8pt} 
\begin{tabular}{cccccccccc}
\toprule
& \multicolumn{3}{c}{Coherence} & \multicolumn{3}{c}{Text-Match} & \multicolumn{3}{c}{Visual Quality} \\
\cmidrule(lr){2-4} \cmidrule(lr){5-7} \cmidrule(lr){8-10}
Method & Bad$\downarrow$ & Normal$\sim$ & Good$\uparrow$ & Bad$\downarrow$ & Normal$\sim$ & Good$\uparrow$ & Bad$\downarrow$ & Normal$\sim$ & Good$\uparrow$ \\
\midrule
AnimateDiff-Lightning & 0.37 & 0.42 & 0.21 & \textbf{0.06} & 0.51 & 0.43 & 0.29 & 0.51 & 0.20 \\
Modelscope T2V & 0.14 & 0.48 & 0.38 & 0.21 & 0.53 & 0.26 & 0.34 & 0.45 & 0.21 \\
Lavie & 0.11 & 0.46 & 0.43 & 0.24 & 0.46 & 0.30 & 0.32 & 0.49 & 0.19 \\
\rowcolor{mylightgray}
\textbf{ConFiner} w/ Lavie & 0.08 & 0.43 & 0.49 & 0.08 & 0.48 & \textbf{0.44} & 0.13 & 0.36 & \textbf{0.51} \\
\rowcolor{mylightgray}
\textbf{ConFiner} w/ Modelscope & \textbf{0.07} & 0.42 & \textbf{0.51} & 0.08 & 0.50 & 0.42 & \textbf{0.09} & 0.41 & 0.50 \\
\bottomrule
\end{tabular}
\vspace{-5pt}
\caption{\textbf{Subjective Evaluation Results.} Each model generates videos using the top 100 prompts from Vbench \cite{b41}. The videos were evaluated by 30 users, with each video being rated as good, normal, or bad on three dimensions.}
\vspace{-10pt}
\label{tab:subjective}
\end{table*}

\subsection{ConFiner-Long Framework}
We also leverage ConFiner to design a pipeline for long video generation. This pipeline generate multiple short video segments and introduces three strategies to ensure consistency and coherence between these segments.

First, we design consistency initialization strategy to promote consistency between segments. The initial noise affects the content of video significantly. To improve the consistency between segments, we first sample a $Noise\_{base}\in \mathbf{R}^{H \times W \times C \times F}$, which is then subjected to frame-wise shuffling to obtain the initial noise for each segment. Sharing base noise enhances the content consistency between segments while shuffling maintains a little randomness.

Although consistency initialization have ensured content consistency between segments, if the motion modes of video structures are not coherent, it will be impossible to combine them into a reasonable long video. Thus, we propose a coherent guidance to promote the motion mode of new segment to follow the preceding segment. In video generation, predicted noises affect the direction of generation and determine the motion mode. So we generate each structure one by one, using noises of the previous segments to guide the subsequent structure. Specifically, during the sampling process, we use the gradient of the L2 loss to guide the sampling direction. The L2 loss is calculated between the predicted noise of the current segment and the noise in the previous segment. The guided noise is calculated as follows:
\begin{equation}
\epsilon_{t}^{S_2} = \hat{\epsilon}_{t}^{S_2} - \gamma \nabla_{\hat{\epsilon}_{t}^{S_2}} \| \hat{\epsilon}_{t}^{S_2} - \epsilon_{t}^{S_1} \|^2
\label{eq18}
\end{equation}
where $\hat{\epsilon}_{t}^{S_2}$ represents the noise of current segment predicted by denoising model at timestep $t$, $\epsilon_{t}^{S_1}$ is the noise of former segment at timestep $t$ and $\gamma$ is a constant.

Additionally, we introduce a staggered refinement mechanism to further improve the overall coherence of the video. In our segmented generation approach, the transition points between segments tend to exhibit the highest inconsistency. Therefore, in long video generation, we perform the Control Stage and Refinement Stage in a staggered manner. Specifically, the latter half of the preceding structure and the former half of the succeeding structure are used as inputs for a same refinement pass. The refinement stage can seamlessly stitch the two structures together, which ensures a more natural and smoother transition between segments.
\begin{equation}
Segment = Refine(Sp_{L/2:L}+Sn_{0:L/2})
\label{eq19}
\end{equation}
Where $Sp$ represents the previous structure, $Sn$ represents the next structure, $L$ represents structures' frames number.

In this way, coherent guidance can make the noise of the two segments similar, which allows the motion mode of the latter segment to inherit that of the previous segment. Additionally, coherence guidance also reduces the pixel distance between noises of two segments, which can help maintain content consistency between segments.

\begin{table*}[t!]
\centering
\renewcommand{\arraystretch}{1}
\small
\setlength{\tabcolsep}{12pt} 
\begin{tabular}{ccccccccc}
\toprule
Method & \shortstack{Control Stage\\ Steps} & $T_e$& \shortstack{Subject \\ Consistency}$\uparrow$ & \shortstack{Motion \\ Smoothness}$\uparrow$ & \shortstack{Aesthetic \\ Quality}$\uparrow$ & \shortstack{Imaging \\ Quality}$\uparrow$ \\ 
\midrule
\rowcolor{mylightgray}
\textbf{ConFiner} w/ Lavie & 4 & 50 & \textbf{0.993} & \textbf{0.991} & 0.703 & 0.733 \\
\rowcolor{mylightgray}
\textbf{ConFiner} w/ Lavie & 4 & 100 & \textbf{0.993} & 0.990 & 0.702 & 0.737 \\
\rowcolor{mylightgray}
\textbf{ConFiner} w/ Lavie & 4 & 200 & 0.992 & 0.989 & \textbf{0.710} & \textbf{0.744} \\
\rowcolor{mylightgray}
\textbf{ConFiner} w/ Lavie & 4 & 300 & 0.978 & 0.986 & 0.383 & 0.303 \\
\rowcolor{mylightgray}
\textbf{ConFiner} w/ Lavie & 4 & 500 & 0.967 & 0.983 & 0.338 & 0.265 \\
\midrule
\textbf{ConFiner} w/ Modelscope & 4 & 50 & \textbf{0.995} & \textbf{0.991} & 0.701 & 0.733 \\
\textbf{ConFiner} w/ Modelscope & 4 & 100 & 0.994 & \textbf{0.991} & 0.698 & 0.733 \\
\textbf{ConFiner} w/ Modelscope & 4 & 200 & 0.994 & 0.990 & \textbf{0.712} & \textbf{0.736} \\
\textbf{ConFiner} w/ Modelscope & 4 & 300 & 0.990 & 0.987 & 0.560 & 0.429 \\
\textbf{ConFiner} w/ Modelscope & 4 & 500 & 0.993 & 0.992 & 0.513 & 0.370 \\
\midrule
\rowcolor{mylightgray}
\textbf{ConFiner} w/ Lavie & 8 & 50 & \textbf{0.994} & \textbf{0.991} & 0.708 & 0.741 \\
\rowcolor{mylightgray}
\textbf{ConFiner} w/ Lavie & 8 & 100 & 0.993 & 0.990 & 0.706 & 0.739 \\
\rowcolor{mylightgray}
\textbf{ConFiner} w/ Lavie & 8 & 200 & 0.991 & 0.989 & 0.716 & 0.742 \\
\rowcolor{mylightgray}
\textbf{ConFiner} w/ Lavie & 8 & 300 & 0.983 & 0.985 & 0.718 & 0.744 \\
\rowcolor{mylightgray}
\textbf{ConFiner} w/ Lavie & 8 & 500 & 0.978 & 0.980 & \textbf{0.721} & \textbf{0.751} \\
\midrule
\textbf{ConFiner} w/ Modelscope & 8 & 50 & \textbf{0.994} & \textbf{0.991} & 0.708 & 0.740 \\
\textbf{ConFiner} w/ Modelscope & 8 & 100 & 0.994 & \textbf{0.991} & 0.707 & 0.739 \\
\textbf{ConFiner} w/ Modelscope & 8 & 200 & 0.993 & 0.990 & 0.716 & 0.742 \\
\textbf{ConFiner} w/ Modelscope & 8 & 300 & 0.992 & 0.989 & 0.720 & 0.747 \\
\textbf{ConFiner} w/ Modelscope & 8 & 500 & 0.991 & 0.987 & \textbf{0.727} & \textbf{0.752} \\
\bottomrule
\end{tabular}
\vspace{-5pt}
\caption{\textbf{Ablation Study of $T_e$.} In most cases, as  $T_e$ increases, the temporal metric decreases and the imaging quality improves. However, when the control stage involves only 4 steps, too high values of \( T_e \) (such as 300 or 500) can lead to imaging collapse.}
\vspace{-5pt}
\label{tab:ablation_t}
\end{table*}

\begin{table*}[t!]
\centering
\renewcommand{\arraystretch}{1}
\small
\setlength{\tabcolsep}{20pt} 
\begin{tabular}{ccccc}
\toprule
\textbf{Time Cost} & \textbf{ConFiner} & \textbf{Lavie\cite{b12}} & \textbf{Animate Diffusion\cite{b15}} & \textbf{Modelscope\cite{b10}} \\ 
\midrule
\rowcolor{mylightgray}
Training & \textit{$0$} & \textit{$>100 \times$ A100 day} & \textit{$>100 \times$ A100 day} & \textit{$>100 \times$ A100 day} \\
Inference & \textit{$\approx$5S} & \textit{$>$1min} & \textit{$>$1min} & \textit{$>$1min} \\
\bottomrule
\end{tabular}
\vspace{-5pt}
\caption{\textbf{Comparison of Training and Inference Time.} We don't need training and only require less than 10\% inference overhead.}
\vspace{-5pt}
\label{tab:timecost}
\end{table*}

\section{Experiments}
In the experiment, we selected AnimateDiff-Lightning \cite{b40} as control expert, and Stable Diffusion 1.5 \cite{b4} as the spatial expert. For the temporal expert, we opted for two open-source models, lavie \cite{b12} and modelscope \cite{b10}.

\subsection{Objective Evaluation}
For objective evaluation experiments, we utilized the cutting-edge benchmark, Vbench \cite{b41}. Vbench provides 800 prompts that test various capabilities of video generation models. In our experiments, each model generated 800 videos using these prompts, and the resulting videos were assessed using four metrics to evaluate their Temporal Quality and Frame-wise Quality.

For Temporal Quality Metrics, we use Subject Consistency and Motion Smoothness. For Frame-wise Quality Metrics, we use Aesthetic Quality and Imaging Quality.

In this experiment, we employed AnimateDiff-Lightning, Lavie, and mocelscope T2V to generate over total timesteps of 10, 20, 50, and 100. We then utilize our ConFiner to conduct generation with 9(4+5) and 18(8+10) timesteps, where $T_e$ is set to 100. All evaluation results are presented in \cref{tab:objective}. Each individual experiment can be completed in 3-5 hours on a single RTX 4090. In each experiment, we repeated for five times with different random seeds. 

\begin{table*}[t!]
\centering
\renewcommand{\arraystretch}{1}
\small
\setlength{\tabcolsep}{7pt} 
\begin{tabular}{ccccccc}
\toprule
Method & \shortstack{Inference\\ Steps} & \shortstack{Denoising\\ Type} & \shortstack{Subject \\ Consistency}$\uparrow$ & \shortstack{Motion \\ Smoothness}$\uparrow$ & \shortstack{Aesthetic \\ Quality}$\uparrow$ & \shortstack{Imaging \\ Quality}$\uparrow$ \\ 
\midrule
\rowcolor{mylightgray}
\textbf{ConFiner} w/ Lavie & 9 & \textbf{Coordinated Denoising} & 0.993 & 0.991 & 0.699 & 0.734 \\
\textbf{ConFiner} w/ Lavie & 9 & Only Temporal Expert & 0.994 & 0.993 & 0.552 & 0.618 \\
\textbf{ConFiner} w/ Lavie & 9 & Only Spatial Expert & 0.883 & 0.907 & 0.749 & 0.766 \\
\rowcolor{mylightgray}
\textbf{ConFiner} w/ Lavie & 18 & \textbf{Coordinated Denoising} & 0.993 & 0.990 & 0.703 & 0.739 \\
\textbf{ConFiner} w/ Lavie & 18 & Only Temporal Expert & 0.993 & 0.991 & 0.583 & 0.632 \\
\textbf{ConFiner} w/ Lavie & 18 & Only Spatial Expert & 0.859 & 0.880 & 0.754 & 0.758 \\
\rowcolor{mylightgray}
\textbf{ConFiner} w/ Modelscope & 9 & \textbf{Coordinated Denoising} & 0.994 & 0.991 & 0.698 & 0.731 \\
\textbf{ConFiner} w/ Modelscope & 9 & Only Temporal Expert & 0.995 & 0.993 & 0.518 & 0.599 \\
\textbf{ConFiner} w/ Modelscope & 9 & Only Spatial Expert & 0.912 & 0.922 & 0.732 & 0.758 \\
\rowcolor{mylightgray}
\textbf{ConFiner} w/ Modelscope & 18 & \textbf{Coordinated Denoising} & 0.994 & 0.991 & 0.707 & 0.739 \\
\textbf{ConFiner} w/ Modelscope & 18 & Only Temporal Expert & 0.993 & 0.992 & 0.577 & 0.641 \\
\textbf{ConFiner} w/ Modelscope & 18 & Only Spatial Expert & 0.861 & 0.893 & 0.765 & 0.772 \\
\bottomrule
\end{tabular}
\vspace{-5pt}
\caption{\textbf{Ablation Study of Denoising Type.} Coordinated denoising achieves a balance between spatial quality and temporal quality.}
\vspace{-5pt}
\label{tab:ablation_d}
\end{table*}

\begin{figure*}[t!]
  \centering
  \includegraphics[width=\textwidth]{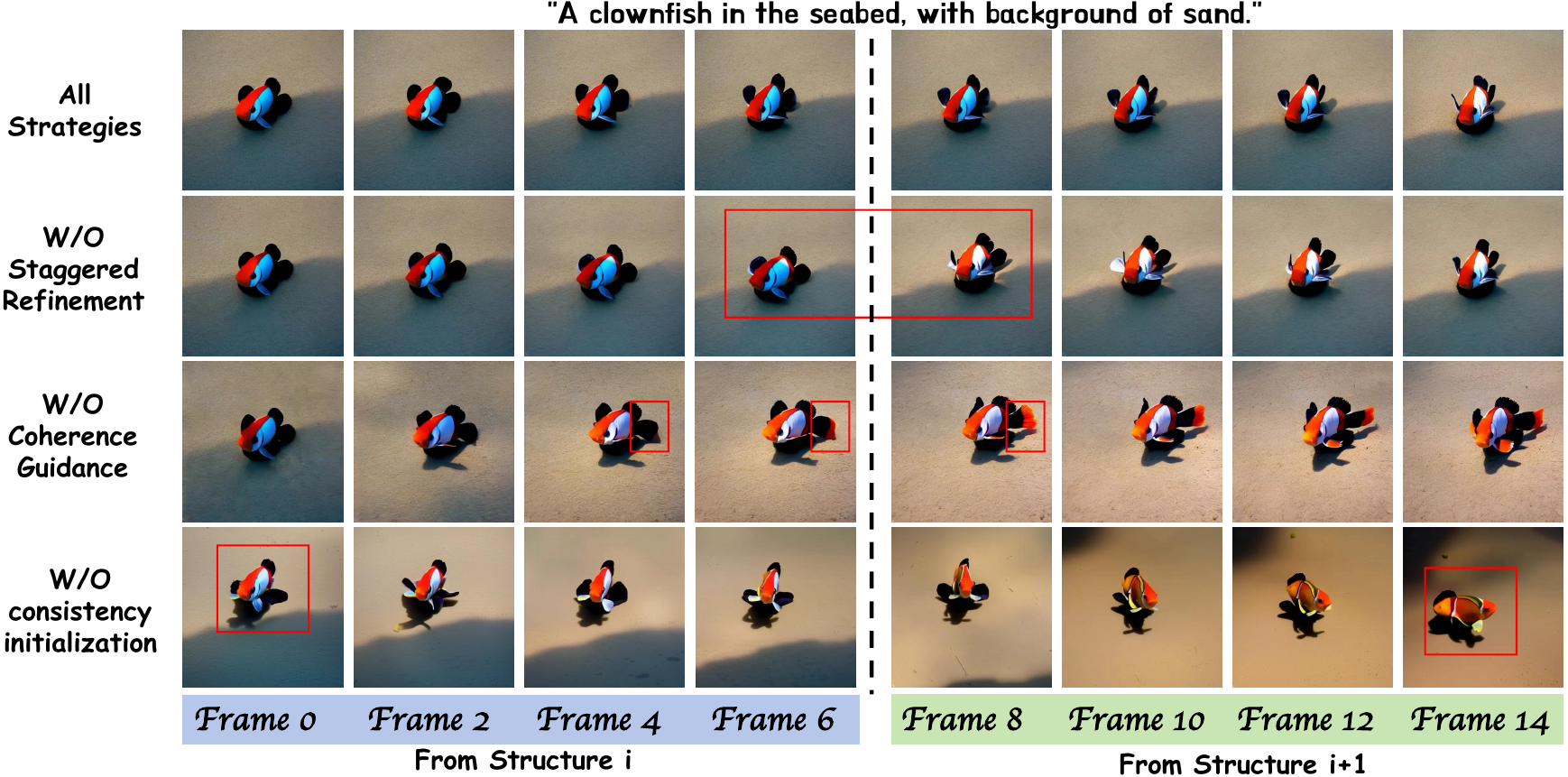}
  \vspace{-5pt}
  \caption{\textbf{Ablation Study on Three Strategies of ConFiner-Long.} Three strategies work together to achieve coherence between segments.}
  \vspace{-10pt}
  \label{ablation_fig}
\end{figure*}

\subsection{Subjective Evaluation}
In our subjective evaluation, we employed our ConFiner with 18 inference steps to generate videos using the top 100 prompts from Vbench. These videos were evaluated alongside those generated by AnimateDiff-Lightning, Modelscope T2V, and Lavie with 50-step inference, by 30 users. Users rated each video across three dimensions: coherence, text-match, and visual quality, each dimension being categorized into three levels: good, normal, and bad. The scoring results are shown in \cref{tab:subjective}.

\subsection{Comparison of Computation Efficiency}
In this section, we compare the training and inference cost of our ConFiner with other video diffusion models. The results are shown in \cref{tab:timecost}.

\subsection{Ablation Study on Control and Refinement Level}
As \cref{eq7}, we apply noise for $T_e$ steps to the videos generated during the control stage to create optimization space for the refinement stage. A larger $T_e$ value increases the impact of the refinement stage. For the four settings same as objective experiment, we set $T_e$ to 50, 100, 200, 300, and 500, with other experimental settings consistent. The performance comparison is shown in \cref{tab:ablation_t}.


\subsection{Ablation Study on Coordinated Denoising}
To verify the effectiveness of coordinated denoising, we conducted ablation experiments on the denoising type during the refinement stage. Specifically, in this experiment, we used Lavie and ModelScope as the temporal experts, setting the total inference steps to 9 and 18, respectively, thus constructing four experimental settings. For each setting, we refined using three different denoising types during the refinement stage: using coordinated denoising; using only the temporal expert; and using only the spatial expert. The performance of the three denoising types is shown in \cref{tab:ablation_d}.

\subsection{Ablation Study on Strategies of ConFiner-Long}
In this section, we conducted ablation experiments on three strategies of ConFiner-Long framework. Using the same preceding video segments, we generated subsequent video segments with either all strategies or only two. The visual comparison of the four video segments against the preceding one is shown in \cref{ablation_fig}. The overall visual comparison between ConFiner-Long and the existing training-free long video generation method Freenoise\cite{b55} is shown in \cref{comparison}.

\section{Conclusion}
In this paper, we introduce ConFiner, a training-free framework that can generate high-quality videos with chain of diffusion model experts. It decouples video generation into three subtasks: structure control, spatial refinement and temporal refinement. Each subtask is handled by a off-the-shelf expert skilled at this task. Additionally, we propose coordinated denoising to enable two expert cooperate at the timestep level when denoising. Based on ConFiner, we also design ConFiner-Long framework to generate long coherent videos by harmonizing various segments. Experimental results confirm that our ConFiner enhances the aesthetics and coherence of generated videos while reducing sampling time significantly. And our ConFiner-Long can generate consistent and coherent videos with up to 600 frames. Our approach paves the way for cost-effective new possibilities in filmmaking, animation production, and video editing.

{
    \small
    \bibliographystyle{ieeenat_fullname}
    \bibliography{main}
}


\end{document}